%% file: main.tex
\documentclass[11pt,logo]{nvidiatechreport}

\usepackage[authoryear,round]{natbib}
\usepackage{array,multirow}
\usepackage{xurl}
\usepackage{placeins}
\input{math_commands}

\hypersetup{
  hidelinks,
  pdftitle={FC-SWE: Failure-Conditioned RL for Long-Horizon Software Engineering Agents},
  pdfauthor={Jia Liufu, Bin Hu, Linglin Jing, Terry Kong, Yuki Huang, Ashwath Aithal, Wenming Yang, Jun Yang}
}
\renewcommand{\today}{2026-09-29}
\title{FC-SWE: Failure-Conditioned RL for Long-Horizon Software Engineering Agents}
\author[1,2]{\mbox{Jia Liufu}}
\author[1]{\mbox{Bin Hu}}
\author[1]{\mbox{Linglin Jing}}
\author[1]{\mbox{Terry Kong}}
\author[1]{\mbox{Yuki Huang}}
\author[1]{\mbox{Ashwath Aithal}}
\author[2]{\mbox{Wenming Yang}}
\author[1]{\mbox{Jun Yang}}
\affil[1]{NVIDIA}
\affil[2]{Tsinghua University}

\newcommand{\method}{\textnormal{\textsc{FC-SWE}}}

\begin{abstract}
\textbf{Abstract:} Repository-level software engineering (SWE) is a challenging
long-horizon setting: agents must reason over extended interactions,
use tools across multiple turns, and adapt to observations from
stateful environments.
Recent work trains SWE agents with reinforcement learning methods
such as Group Relative Policy Optimization (GRPO), which independently
sample multiple trajectories per issue, test the resulting patches,
and compare terminal rewards within a fixed group.
However, this training setup does not reuse verifier feedback from
failed patches as context for subsequent attempts, even though this
feedback contains valuable diagnostic information about what went wrong.
Training on recovery trajectories is challenging because the preceding
outcome determines whether the next trajectory is generated, while
the failed execution determines its conditioning context.
We introduce FC-SWE, a failure-conditioned RL framework that incorporates
recovery attempts into policy training.
After a patch fails verification, FC-SWE restores the repository to its
original task state and uses the failed patch and verifier feedback
as context for a recovery trajectory.
FC-SWE adapts GRPO to these chains of complete, multi-turn tool-use
trajectories through two mechanisms.
Trajectory-local rewards preserve each attempt's verifier outcome,
preventing recovery success from rewarding an earlier failed patch.
Active-set advantage estimation forms a comparison group from all
initial and recovery trajectories actually executed for the same issue,
so failed attempts remain in the group while unexecuted attempts
are excluded.
On all 500 SWE-bench Verified tasks under a verifier-assisted protocol,
FC-SWE with Qwen3.5-4B and SWE-agent achieves 41.7\% Resolved@1 and
52.8\% Resolved@2, compared with 38.9\% and 48.5\% for GRPO.
Although trained with at most two attempts per chain, FC-SWE reaches
70.7\% Resolved@11 under an eleven-attempt test-time budget.
\end{abstract}

\begin{document}
\maketitle

\input{sections/01_introduction}
\input{sections/02_related_work}
\input{sections/03_method}
\input{sections/04_experiments}
\input{sections/07_conclusion}

\subsection*{AI use statement}
Generative AI tools were used to refine hypotheses and manuscript structure, provide feedback on methodology and experimental design, interpret author-provided results, discover and summarize related literature, and draft and edit manuscript text. The authors checked AI-assisted claims against the training implementation, experiment logs, primary literature, and original evaluation records. No AI-generated value is presented as a measured experimental result. The authors take responsibility for the final content of this work, including text, claims, and artifacts produced with the aid of generative AI.

\FloatBarrier
\bibliography{references}
\bibliographystyle{plainnat}

\clearpage
\appendix
\input{appendix}

\end{document}

%% file: math_commands.tex
\usepackage{amsmath,amsfonts,bm}

\def\eqref#1{equation~\ref{#1}}

\def\1{\bm{1}}

\DeclareMathAlphabet{\mathsfit}{\encodingdefault}{\sfdefault}{m}{sl}
\SetMathAlphabet{\mathsfit}{bold}{\encodingdefault}{\sfdefault}{bx}{n}



%% file: sections/01_introduction.tex
\section{Introduction}
\label{sec:introduction}

Large language models (LLMs) can generate code from natural-language
instructions. Coding agents extend this capability by using tools
to inspect files, edit code, and run tests, with execution feedback
informing subsequent actions~\citep{yang2024sweagent,wang2024openhands}.
These capabilities support repository-level software engineering
(SWE), where agents must resolve issues in existing codebases rather
than generate standalone programs~\citep{jimenez2023swebench}.
Each action can change the repository or reveal information needed
for the next decision~\citep{yao2023react,yang2024sweagent}.
Solving such tasks therefore requires cycles of inspection,
editing, execution, and revision across repeated interactions with
the environment.

\begin{figure}[t]
    \centering
    \includegraphics[
        width=0.96\linewidth,
        keepaspectratio
    ]{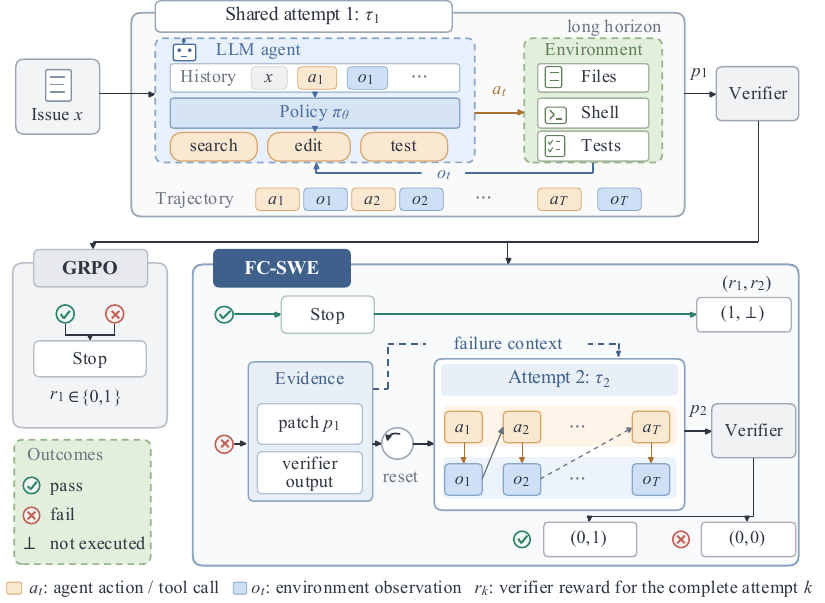}
    \caption{
    Comparison of trajectory execution in GRPO and \method{}.
    During training, our GRPO baseline samples independent
    attempts, whereas \method{} launches a recovery trajectory
    after failure, conditioned on the failed patch and
    verifier output.
    Each executed trajectory retains its own verifier reward.
    }
    \label{fig:method-overview}
\end{figure}

To improve software reasoning and agent behavior over these
extended interactions, recent work has applied reinforcement
learning (RL) to repository-level
SWE~\citep{wei2025swerl,golubev2025longcontext,deepswe2025}.
In execution-based pipelines, the sequence of model outputs,
tool calls, and environment observations forms a complete
trajectory that culminates in a candidate code patch.
A test-suite verifier evaluates the patch, and its outcome
is used as a terminal reward for policy
optimization~\citep{golubev2025longcontext,deepswe2025}.
Although effective, training only on independent attempts
does not explicitly teach the policy to use the diagnostic
information from a failed verification in a subsequent
recovery attempt.

A failed verification, however, provides more than a negative
outcome: failing tests, error messages, and tracebacks can
reveal why the patch was rejected and provide concrete
guidance for recovery~\citep{gehring2024rlef}.
This creates an opportunity to use not only the failed
trajectory's terminal outcome, but also its diagnostic
evidence to guide subsequent
attempts~\citep{shinn2023reflexion,gehring2024rlef}.
Incorporating this recovery process into reinforcement
learning, however, is not straightforward.
A recovery trajectory is generated only after a preceding
trajectory fails, and its context is constructed from the
failed patch and verifier feedback.
Preceding executions therefore determine both which recovery
trajectories enter training and the contexts under which
they are generated, requiring explicit decisions about how
verifier rewards are assigned and which trajectories form
the comparison groups for advantage estimation.

To address these challenges, we introduce FC-SWE, a
failure-conditioned reinforcement learning framework that
incorporates post-failure recovery into policy optimization
for long-horizon SWE agents.
As illustrated in Figure~\ref{fig:method-overview}, after a
submitted patch fails verification, FC-SWE restores the
repository to its original task state, provides the policy
with the failed patch and verifier feedback alongside the
original issue, and generates a recovery trajectory.
This process forms an outcome-dependent recovery chain:
the initial trajectory is always generated, whereas each
recovery trajectory exists only if the preceding trajectory
fails.

FC-SWE adapts Group Relative Policy Optimization
(GRPO)~\citep{shao2024deepseekmath} to these recovery chains
through two mechanisms.
Trajectory-local reward assignment gives each trajectory
the verifier outcome of its own patch, so a chain that
fails and then recovers retains the rewards $[0,1]$;
the later success does not change the earlier failure's label.
Active-set advantage estimation pools the initial and recovery
trajectories actually executed for the same issue and uses
the same leave-one-out reward baseline as our GRPO
implementation~\citep{ahmadian2024back}.
This grouping allows comparisons across attempts while
excluding unexecuted recovery slots from chains that
have already succeeded.

We evaluate on all 500 SWE-bench Verified tasks using SWE-agent.
We compare the initial Qwen3.5-4B policy (Base) with two
policies separately trained from it using GRPO and FC-SWE;
all three use the same feedback-enabled evaluation
controller and attempt budgets.
Averaged over three decoding seeds, FC-SWE achieves
41.7\% Resolved@1 and 52.8\% Resolved@2, compared with
27.9\% and 36.3\% for Base, and 38.9\% and 48.5\% for
GRPO (Table~\ref{tab:main}).
At a matched two-attempt budget, feedback-conditioned recovery
performs worse than independent resampling for Base and GRPO
but better for FC-SWE; context ablations further show that
verifier evidence improves recovery beyond patch-only
conditioning (Section~\ref{sec:resample}).
Although trained with at most two attempts per chain,
the same FC-SWE policy reaches 70.7\% cumulative resolution
with up to eleven verifier-guided attempts at test time,
compared with 67.3\% for GRPO
(Section~\ref{sec:beyond}).

Our main contributions are:

\begin{itemize}
    \item We introduce FC-SWE, a failure-conditioned RL
    framework for long-horizon SWE agents that extends policy
    training beyond failed verification by restoring the
    repository to its original task state and generating
    recovery trajectories conditioned on failed patches
    and verifier feedback.

    \item To optimize these outcome-dependent recovery
    chains, we introduce two complementary mechanisms.
    Trajectory-local reward assignment preserves each
    trajectory's own verifier outcome, preventing a later
    recovery success from changing the reward assigned to
    an earlier failure.
    Active-set advantage estimation compares the initial
    and recovery trajectories actually executed for each
    task while excluding unexecuted recovery slots from
    group statistics.

    \item On all 500 SWE-bench Verified tasks, FC-SWE improves
    over the GRPO baseline trained without recovery
    trajectories by 2.9 and 4.3 percentage points on
    Resolved@1 and Resolved@2, respectively.
    Controlled evaluations further show that FC-SWE benefits
    from verifier evidence and outperforms independent
    resampling under the same two-attempt budget.
    Despite training with at most two attempts per chain,
    FC-SWE remains ahead of GRPO when evaluated at test time
    with up to eleven verifier-guided attempts.
\end{itemize}

%% file: sections/02_related_work.tex
\section{Related Work}
\label{sec:related}

\paragraph{Software-engineering agents and RL.}
SWE-bench evaluates repository-level issue resolution with executable
tests~\citep{jimenez2023swebench}, while SWE-agent studies how
agent--computer interfaces affect performance~\citep{yang2024sweagent}.
OpenHands, SWE-Gym, and R2E-Gym provide agent infrastructure, training
environments, and verifiers~\citep{wang2024openhands,pan2024swegym,jain2025r2egym}.
SWE-RL learns from software evolution data~\citep{wei2025swerl},
while other work applies RL to long-context agent
trajectories~\citep{golubev2025longcontext,deepswe2025}.
Satori-SWE trains agentless iterative patch refinement for test-time
scaling using reward differences between successive
patches~\citep{zeng2025satoriswe}.
FC-SWE trains recovery after patch verification through complete
tool-use trajectories.

\paragraph{Group-relative and long-horizon agent RL.}
GRPO uses critic-free group-relative optimization~\citep{shao2024deepseekmath};
DAPO targets stable large-scale reasoning RL~\citep{yu2025dapo}.
GiGPO combines episode-level comparisons with groups of steps sharing
environment states~\citep{feng2025gigpo}, Agent Lightning decomposes
agent executions into learning transitions~\citep{luo2025agentlightning},
and CompactionRL estimates advantages across context-compaction
boundaries~\citep{li2026compactionrl}.
LaMer restarts failed episodes from the original state, carries
reflection forward, and optimizes returns including later
outcomes~\citep{jiang2026lamer}.
PivoARL retries from error states and isolates future credit from
erroneous suffixes~\citep{guo2026pivoarl}.
FC-SWE assigns each complete attempt its own verifier reward and
compares only executed initial and recovery trajectories for the
same issue.

\paragraph{Feedback and iterative refinement.}
Self-Refine and Reflexion use feedback without weight
updates~\citep{madaan2023selfrefine,shinn2023reflexion}, but self-repair
can underperform independent resampling for weaker models at fixed
generation budgets~\citep{olausson2024selfrepair}.
RISE learns revision from multi-turn data~\citep{qu2024rise},
while SCoRe trains two-turn self-correction without external feedback
at inference~\citep{kumar2024score}.
RLEF uses public-test feedback for multi-turn code-synthesis
RL~\citep{gehring2024rlef}, and RefineRL combines a Skeptical-Agent
with self-refinement RL for competitive programming~\citep{fu2026refinerl}.
ContextRL recovers positives from all-negative groups using
reference-solution-conditioned mistake reports, then removes failed
responses and reports from context before optimizing recovered
outputs as standalone samples~\citep{lu2026contextrl}.
MulFeRL uses feedback to regenerate all-failed groups and combines
within-state GRPO with Feedback-Contrastive Optimization across
all-failed to all-solved transitions~\citep{li2026mulferl}.
FC-SWE triggers recovery per failed trajectory within the attempt
budget, restarts from the original repository state, and retains the
failed patch and verifier output as context during
optimization.

%% file: sections/03_method.tex
\section{Method}
\label{sec:method}

\method{} extends GRPO to outcome-dependent recovery chains:
failed attempts trigger new trajectories conditioned on
the failed patch and verifier feedback after a clean
repository reset.
Trajectory-local rewards preserve individual verifier
outcomes, while active-set advantage estimation compares
executed trajectories for the same issue.
Figure~\ref{fig:system} summarizes the training pipeline.

\input{figures/precompiled/includes/system_pipeline}

\subsection{Problem Setup and GRPO Baseline}
\label{sec:setup}
For a software issue $x$, we define an attempt as one complete agent run beginning from the issue's original repository state and ending after its submitted patch is evaluated by the verifier.
Each executed attempt produces one trajectory comprising policy outputs, tool calls, and environment observations.
We use $\tau_c$ to denote the trajectory of group member $c$ and $y_c\in\{0,1\}$ to denote the binary verifier outcome of its submitted patch.
The GRPO baseline independently samples a fixed group of $G$ trajectories $\{\tau_c\}_{c=1}^{G}$ for the same issue.
Our implementation combines GRPO's critic-free group-relative policy update~\citep{shao2024deepseekmath} with a leave-one-out reward baseline~\citep{ahmadian2024back}:
\begin{equation}
b_c^{\mathrm{GRPO}}
=
\frac{1}{G-1}\sum_{d\ne c}y_d,
\qquad
s_{x,-c}
=
\operatorname{Std}\!\left(\{y_d:d\ne c\}\right),
\end{equation}
\begin{equation}
A_c^{\mathrm{GRPO}}
=
\begin{cases}
\dfrac{y_c-b_c^{\mathrm{GRPO}}}{s_{x,-c}+\epsilon},
& s_{x,-c}>0, \\[6pt]
y_c-b_c^{\mathrm{GRPO}},
& s_{x,-c}=0,
\end{cases}
\label{eq:grpo}
\end{equation}

where $b_c^{\mathrm{GRPO}}$ and $s_{x,-c}$ are respectively
the mean and sample standard deviation of the other $G-1$
trajectory rewards, and $\epsilon=10^{-6}$ is a
numerical-stability constant.

Unlike \method{}, the baseline samples all $G$ trajectories
independently; a failed patch neither triggers nor conditions
another trajectory in the training group.

\subsection{Outcome-Dependent Recovery Chains}
\label{sec:chains}
For each software issue $x$, each of the $G$ initially sampled trajectories seeds one recovery chain indexed by $c\in\{1,\ldots,G\}$.
Chain $c$ contains at most $R$ attempts, and $\tau_{c,r}$ denotes the trajectory produced by attempt $r\in\{1,\ldots,R\}$ in that chain.
A verifier returns a binary outcome $y_{c,r}\in\{0,1\}$ for every executed trajectory.
Let $q_{c,r}\in\{0,1\}$ indicate whether attempt $r$ in chain $c$ is executed.
The initial attempt is always active, and the activity of later attempts evolves recursively as
\begin{equation}
q_{c,1}=1,
\qquad
q_{c,r}
=
\begin{cases}
1-y_{c,r-1}, & q_{c,r-1}=1,\\
0, & q_{c,r-1}=0,
\end{cases}
\quad r>1.
\label{eq:active}
\end{equation}
Thus, a recovery attempt is executed if and only if the preceding attempt was executed and failed.
Our main training configuration uses $G=8$ and $R=2$.

All $G$ initial trajectories are sampled independently, whereas each recovery trajectory is sampled only after its predecessor fails.
The preceding verifier outcome determines whether the recovery trajectory exists, while the preceding failed trajectory supplies the evidence on which it is conditioned.
Because verifier outcomes depend on policy-generated patches, the policy affects both the trajectory contents and the active set entering the update.

When an executed attempt $r<R$ fails, the controller records its submitted patch and diagnostic verifier output, restores the repository to its original task state, and constructs the context for attempt $r+1$ as
\begin{equation}
h_{c,r+1}
=
\mathcal{C}\!\left(
x,
\mathrm{patch}(\tau_{c,r}),
\mathrm{verify}(\tau_{c,r})
\right).
\label{eq:recoverycontext}
\end{equation}
Here, $h_{c,r+1}$ is the context supplied to the recovery attempt, $\mathrm{patch}(\tau_{c,r})$ extracts the submitted code patch, and $\mathrm{verify}(\tau_{c,r})$ denotes the diagnostic verifier output, distinct from the binary outcome $y_{c,r}$.
The context-construction operator $\mathcal{C}$ formats and truncates the original issue and failure evidence to a fixed context budget.
Restoring the repository prevents transient files, processes, and tool state from leaking across attempts, while the failed patch and verifier output remain available as explicit diagnostic context for recovery.

For static distributed batching, the implementation may allocate all $G\times R$ rows.
Rows with $q_{c,r}=0$ are batching placeholders rather than policy-generated trajectories and are excluded from both reward statistics and policy optimization.

\subsection{Trajectory-Local Verifier Rewards}
\label{sec:rewards}
The first design choice concerns how verifier outcomes are assigned across trajectories in a recovery chain.
Each executed attempt begins with the repository in its original task state, produces one complete trajectory and candidate patch, and receives a separate verifier evaluation.
\method{} assigns each verifier outcome only to the trajectory that produced the corresponding patch:
\begin{equation}
r_{c,r}
=
\begin{cases}
y_{c,r}, & q_{c,r}=1,\\
\bot, & q_{c,r}=0,
\end{cases}
\label{eq:localreward}
\end{equation}
where $r_{c,r}$ is the training reward associated with trajectory $\tau_{c,r}$, and $\bot$ denotes an attempt position that was never executed rather than a trajectory with zero reward.
For $R=2$, the possible reward sequences are $[1,\bot]$ after initial success, $[0,1]$ after successful recovery, and $[0,0]$ after two failures.

One alternative is to broadcast the final chain outcome to every executed trajectory in the chain.
A recovered chain with verifier outcomes $[0,1]$ would then receive training rewards $[1,1]$, incorrectly assigning a positive reward to the rejected initial patch solely because a later policy invocation succeeded.
At the opposite extreme, optimizing only the final executed trajectory would remove the preceding failed trajectories from the policy loss, eliminating the rejected behavior that provides contrast for the successful recovery.
\method{} instead retains every executed trajectory and preserves its own verifier outcome.
A later success therefore does not alter an earlier trajectory's reward label, although it can affect that trajectory's relative advantage through the comparison set defined in the next section.

\subsection{Active-Set Advantage Estimation}
\label{sec:advantages}

We retain the leave-one-out reward baseline~\citep{ahmadian2024back}
and the normalization convention of our GRPO implementation
(Equation~\ref{eq:grpo}).
The change is the comparison group: for each issue $x$, we pool
all executed initial and recovery trajectories,
\begin{equation}
\mathcal{V}_x=\{(c,r):q_{c,r}=1\}.
\label{eq:active-set}
\end{equation}
Trajectories are grouped by the original issue, even when their
failure-conditioned contexts differ. For $i=(c,r)\in\mathcal{V}_x$
with reward $r_i=r_{c,r}$, the baseline is
\begin{equation}
b_i=\frac{1}{|\mathcal{V}_x|-1}
\sum_{\substack{j\in\mathcal{V}_x\\j\ne i}}r_j.
\label{eq:advantage}
\end{equation}
We obtain $A_i$ by normalizing $r_i-b_i$ using the sample standard
deviation of these same comparison rewards, with the
zero-variance fallback in Equation~\ref{eq:grpo}.
Inactive rows enter neither the reward statistics nor the loss.

For a recovered chain with rewards $[0,1]$, this grouping gives
the successful recovery a positive advantage and the preceding
failure a negative advantage, without changing either reward.
It excludes padding while preserving cross-attempt comparisons
that separate attempt-index groups would omit.
Because recovery trajectories exist only after failure,
the comparison set is outcome-dependent; this update does not
inherit the standard unbiasedness guarantee of an
action-independent baseline.

\subsection{Policy Optimization}
\label{sec:optimization}

Trajectory advantages are broadcast to valid policy-generated
tokens and optimized with a clipped policy-gradient
objective~\citep{schulman2017ppo}, normalized by the total
number of valid generated tokens across the distributed batch.
Prompts, environment observations, copied failure context,
and padding receive zero loss weight.
A failed patch is loss-bearing only in the trajectory that
generated it, not when copied into a recovery prompt.
For assistant messages flagged as invalid tool calls or
malformed thinking, token advantages are overwritten with $-5$.
These runtime-local penalties and optimization settings are
shared by GRPO and \method{}, except where explicitly ablated.

%% file: figures/precompiled/includes/system_pipeline.tex
\begin{figure}[!t]
    \centering
    \includegraphics[
        width=0.96\linewidth
    ]{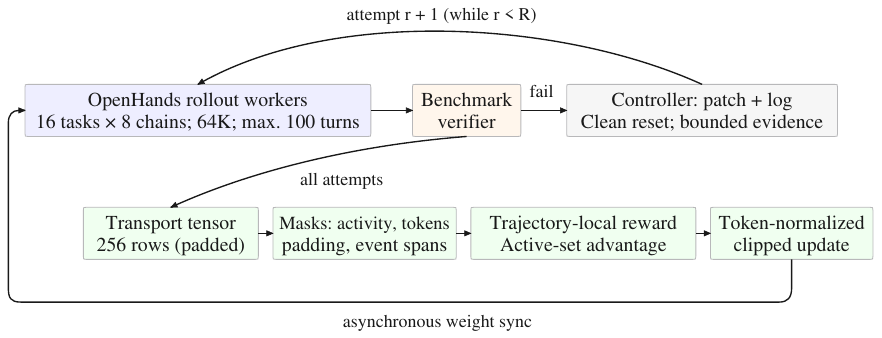}
    \caption{
    End-to-end FC-SWE training system.
    Rollout workers execute verifier-delimited attempts; failed
    chains are reset and resumed with bounded failure evidence.
    Executed trajectories are packed with inactive transport padding,
    assigned trajectory-local rewards and active-set advantages,
    and optimized with a token-normalized policy update.
    }
    \label{fig:system}
\end{figure}

%% file: sections/04_experiments.tex
\section{Experiments}
\label{sec:experiments}

\subsection{Experimental Setup}
\label{sec:experimental-setup}

\paragraph{Evaluation.}
We evaluate on all 500 SWE-bench Verified
tasks~\citep{jimenez2023swebench} using
SWE-agent~\citep{yang2024sweagent}.
The controlled Qwen3.5-4B comparison uses the same
feedback-enabled controller and attempt budgets, with a
64K-token context and at most 100 model--tool interactions
per attempt. Execution stops after success or budget exhaustion.
Unless otherwise specified, retries use a clean reset and
failure context as defined in Section~\ref{sec:chains}.
Verifier choice is discussed in Appendix~\ref{app:verifier}.
We report first-attempt and cumulative verifier-assisted
success, together with conditional recovery
(Table~\ref{tab:main}), rather than independent-sampling Pass@$K$.

\paragraph{Training and baselines.}
Training uses 4{,}518 tasks from ten R2E-Gym
repositories~\citep{jain2025r2egym}, with rollouts generated
using OpenHands~\citep{wang2024openhands}.
We use the initial Qwen3.5-4B policy as Base.
Starting from this same policy, we separately train GRPO
and \method{} for the controlled comparison.
Both RL methods sample 16 tasks with eight initial trajectories
per task at each optimization step; \method{} additionally
generates one recovery trajectory after each initial failure.
Qwen3.5-9B and Nemotron-3-Nano-4B provide cross-model reference
results. Unless stated otherwise, benchmark results use three decoding seeds,
with aggregation defined in Table~\ref{tab:main}.
Hyperparameters and checkpoint-selection details appear in
Appendix~\ref{app:details}, per-seed
results in Appendix~\ref{app:scaling-details}, and cross-model
runtime settings in Appendix~\ref{app:execution-diagnostics}.

\subsection{FC-SWE Improves Initial Solving and Recovery}
\label{sec:mainresults}

Table~\ref{tab:main} reports results for the evaluated models and
the two policies trained from Qwen3.5-4B. Within this same-model
comparison, GRPO raises first-attempt resolution from 27.9\% to 38.9\%.
\method{} further improves $\mathrm{Resolved@1}$ to 41.7\% and
$\mathrm{Resolved@2}$ to 52.8\%. Among initial failures, its conditional
recovery reaches 19.0\%, compared with 15.7\% for GRPO and 11.6\% for
Base. Thus, the recovery-training configuration improves benchmark
recovery without sacrificing first-attempt performance.

\begin{table}[t]
\caption{Results on all 500 SWE-bench Verified tasks.
All results are evaluated by us.
Resolved@1 and Resolved@2 are first-attempt and cumulative
two-attempt success, averaged over three decoding seeds (avg@3).
Conditional recovery pools second-attempt successes over
initial failures across seeds. Retries use failure feedback.
Higher is better.}
\label{tab:main}
\centering
\small
\setlength{\tabcolsep}{4pt}
\renewcommand{\arraystretch}{1.12}
\begin{tabular*}{\linewidth}{@{\extracolsep{\fill}}lrrr@{}}
\toprule
Model / policy & Resolved@1 & Resolved@2 & Conditional recovery \\
\midrule
Qwen3.5-4B~\citep{qwen35_4b_model} & 27.9\% & 36.3\% & 11.6\% \\
Qwen3.5-9B~\citep{qwen35_9b_model} & 41.2\% & 50.4\% & 15.6\% \\
Nemotron-3-Nano-4B~\citep{nemotron3_nano4b_model} & 28.2\% & 35.8\% & 10.6\% \\
\midrule
\multicolumn{4}{@{}l}{\textit{Qwen3.5-4B RL training}} \\
\addlinespace[2pt]
GRPO~\citep{shao2024deepseekmath} & 38.9\% & 48.5\% & 15.7\% \\
\textbf{\method{}} (ours)
& \textbf{41.7\%} & \textbf{52.8\%} & \textbf{19.0\%} \\
\bottomrule
\end{tabular*}
\end{table}

\begin{figure}[t]
\centering
\includegraphics[width=\linewidth]{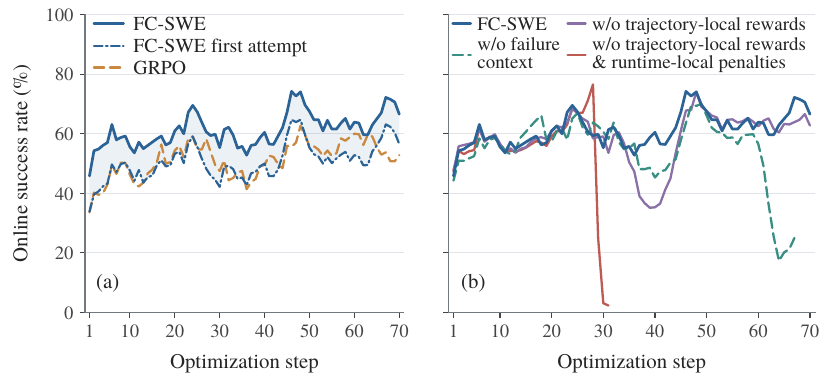}
\caption{Online training success during optimization.
(a) First-attempt and cumulative two-attempt success
of \method{}, compared with single-attempt GRPO;
shading shows the gain from recovery.
(b) Ablations of failure context and reward design in \method{}.
Curves use five-step moving averages, except for the collapse segment.}
\label{fig:training-curve}
\end{figure}

Figure~\ref{fig:training-curve}(a) shows that second-attempt recovery
contributes substantially to online training success. Over the first
70 optimization steps, \method{} and GRPO have similar first-attempt
success rates on average, with a mean difference of $-0.7$ percentage
points for \method{}. The second attempt adds 10.6 percentage points
to \method{}'s success rate, yielding a 9.9-point average advantage
over GRPO's single-attempt rate. This online comparison includes
the additional recovery attempt; Table~\ref{tab:main} instead compares
the trained policies under matched evaluation attempt budgets.

\subsection{Ablations of FC-SWE Training}
\label{sec:dynamics}

We examine three training ablations through online dynamics
(Figure~\ref{fig:training-curve}(b)) and held-out evaluation
(Table~\ref{tab:ablations}). The two reward/penalty variants are
continued from the same step-26 checkpoint; their comparison concerns
the subsequent training segment. Inference-time context controls are
analyzed separately in Section~\ref{sec:resample}.

\textbf{Failure context improves recovery beyond additional retries.}
The \emph{w/o failure context} variant (Fresh-retry) retains
failure-triggered second attempts, clean resets, trajectory-local
rewards, and active-set advantage estimation. Its second training
attempt receives only the original issue, without the failed patch,
verifier feedback, or prior interaction history. Both trained policies
are evaluated with the same feedback-enabled controller.
Removing training-time failure context reduces $\mathrm{Resolved@1}$
from 41.7\% to 40.4\%, $\mathrm{Resolved@2}$ from 52.8\% to 47.9\%,
and conditional recovery from 19.0\% to 12.6\%.
The shared retry mechanism alone therefore does not recover the full
benefit of failure-conditioned training.

\textbf{Trajectory-local rewards improve downstream recovery.}
The \emph{w/o trajectory-local rewards} variant assigns the final
chain outcome to every executed trajectory: a recovered chain with
verifier outcomes $[0,1]$ receives training rewards $[1,1]$.
Its online success undergoes a mid-training drop before recovering.
At the selected checkpoints, evaluated with one decoding seed,
conditional recovery decreases from 20.4\% to 17.7\%, while
$\mathrm{Resolved@2}$ decreases from 54.8\% to 51.8\%
(Table~\ref{tab:ablations}, block B).
These results support preserving the distinction between failed and
successful attempts during reward assignment.

\textbf{Removing runtime-local penalties destabilizes shared-reward training.}
We additionally remove the penalties on invalid tool calls and
malformed thinking from the shared-reward variant.
In the post-checkpoint segment, online success falls from 76.6\%
at step 28 to 2.3\% at step 31
(Figure~\ref{fig:training-curve}(b)); this run was not evaluated
downstream. This contrast supports the stabilizing role of these
penalties within the shared-reward configuration, rather than
isolating their effect under every reward scheme.

\begin{table}[t]
\caption{Training ablations with failure context retained
at evaluation.
Block A uses three decoding seeds; Block B uses one.
Metric definitions and aggregation follow Table~\ref{tab:main}.}
\label{tab:ablations}
\centering
\small
\setlength{\tabcolsep}{4pt}
\renewcommand{\arraystretch}{1.10}
\begin{tabular*}{\linewidth}{@{\extracolsep{\fill}}lrrr@{}}
\toprule
Training variant
& Resolved@1
& Resolved@2
& Conditional recovery \\
\midrule
\multicolumn{4}{@{}l}{
\textit{A. Training-time failure context}
} \\
\method{}
& \textbf{41.7\%}
& \textbf{52.8\%}
& \textbf{19.0\%} \\
w/o failure context
& 40.4\% & 47.9\% & 12.6\% \\
\midrule
\multicolumn{4}{@{}l}{
\textit{B. Reward assignment and runtime penalties}
} \\
\method{}
& \textbf{43.2\%}
& \textbf{54.8\%}
& \textbf{20.4\%} \\
w/o trajectory-local rewards
& 41.4\% & 51.8\% & 17.7\% \\
\shortstack[l]{w/o trajectory-local rewards\\
\& runtime-local penalties}
& \multicolumn{3}{c@{}}{
\textit{Not evaluated (training collapse)}
} \\
\bottomrule
\end{tabular*}
\end{table}
Additional tool-use, generation-length, and logged-truncation
statistics are reported in Appendix~\ref{app:execution-diagnostics}.
These describe training behavior; inference efficiency is evaluated
separately in Section~\ref{sec:beyond}.

\subsection{Learning to Use Failure Evidence}
\label{sec:resample}
\label{sec:pending}

\begin{figure}[t]
\centering
\includegraphics[width=\linewidth]{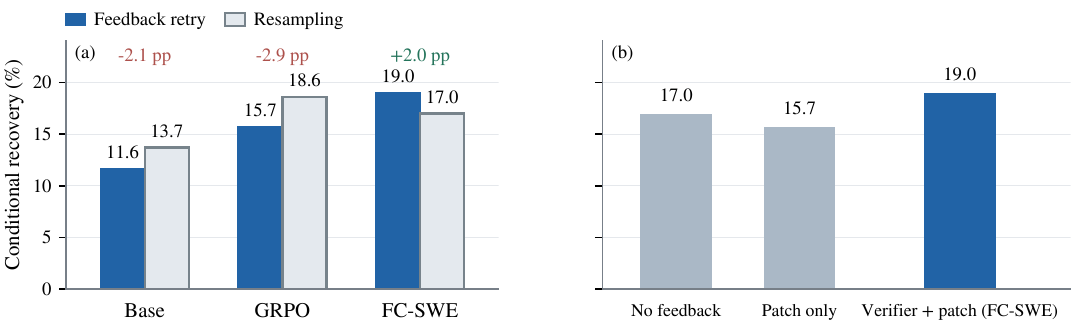}
\caption{Failure-evidence ablations.
(a) Feedback retries versus resampling at a matched
two-attempt budget; labels show differences in percentage points.
(b) Second-attempt context ablations using the final
\method{} policy.}
\label{fig:feedback-mechanism}
\end{figure}

An additional attempt can succeed through resampling alone.
We therefore compare feedback-conditioned retries with independent
resampling under the same policy, controller, decoding configuration,
and two-attempt budget (Figure~\ref{fig:feedback-mechanism}(a)).
For Base and GRPO, feedback-conditioned retries yield lower conditional
recovery than resampling: 11.6\% versus 13.7\%, and 15.7\% versus 18.6\%,
respectively. \method{} reverses this pattern, achieving 19.0\% with
feedback versus 17.0\% with resampling. Thus, providing failure context
does not uniformly improve recovery; a positive feedback advantage
emerges for the policy trained to recover from failed attempts.

We next vary the second-attempt context while holding the \method{}
policy, controller, and initial-failure sets fixed
(Figure~\ref{fig:feedback-mechanism}(b)). Providing only the failed patch
yields 15.7\% conditional recovery, compared with 17.0\% without failure
context. Adding verifier evidence to the patch raises recovery to
19.0\%, a 3.3-percentage-point improvement over patch-only conditioning.
This supports the value of verifier evidence beyond simply showing
the previous patch. Together with the training-time context ablation,
these results distinguish the value of learning from failure context
from the benefit of merely generating another attempt. Additional
verifier-only and shuffled-evidence controls appear in
Appendix~\ref{app:ablations}.

\subsection{Test-Time Scaling and Efficiency}
\label{sec:beyond}

\begin{figure}[t]
\centering
\includegraphics[width=\linewidth]{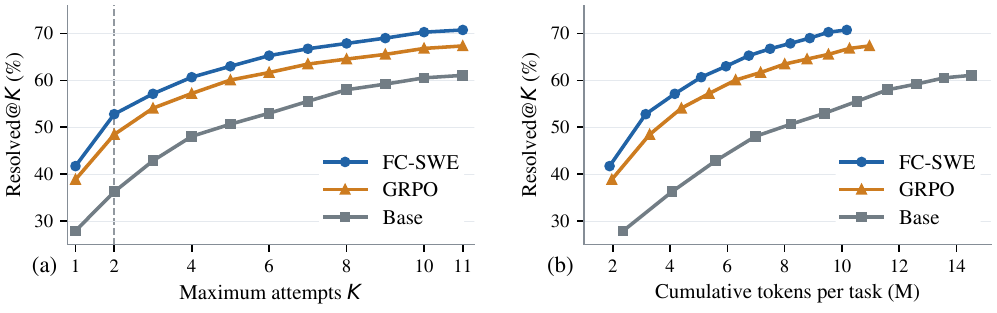}
\caption{Test-time scaling and token efficiency
(three-seed means).
(a) Resolved@$K$ versus attempt budget; the dashed line
marks the two-attempt training horizon.
(b) Resolved@$K$ versus estimated cumulative tokens per task.
Token accounting is detailed in
Appendix~\ref{app:scaling-details}.}
\label{fig:scaling-efficiency}
\end{figure}

\paragraph{Scaling beyond the training horizon.}
Although \method{} is trained with at most two attempts, its recovery
behavior extends to larger test-time budgets. With the same controller
and stop-on-success rule, cumulative resolution increases from 52.8\%
at $K=2$ to 70.7\% at $K=11$, compared with 67.3\% for GRPO and 61.1\%
for Base (Figure~\ref{fig:scaling-efficiency}(a)). The gains diminish
at larger budgets, but the advantage persists beyond the training
horizon. Full per-seed counts appear in Table~\ref{tab:scaling-full}.

\paragraph{Inference token cost.}
Figure~\ref{fig:scaling-efficiency}(b) plots cumulative resolution
against estimated tokens per task, including failed attempts and
stopping after success. Costs combine recorded input tokens and
estimated generated tokens (Appendix~\ref{app:scaling-details}).
At the lowest-cost evaluated budgets reaching at least 50\% resolution,
\method{}, GRPO, and Base use approximately 3.15M, 4.39M, and 8.22M
tokens per task; at 60\%, they use 5.08M, 6.29M, and 13.57M.
These discrete, non-interpolated points can exceed the targets by
different amounts. The comparison concerns inference token cost,
not training compute or measured wall-clock speed.

%% file: sections/07_conclusion.tex
\section{Conclusion}

FC-SWE trains SWE agents to recover from failed patches using
trajectory-local rewards and active-set advantage estimation.
On Qwen3.5-4B, it improves initial solving and recovery over GRPO
and benefits from verifier evidence at a matched two-attempt budget.
Trained with at most two attempts, it reaches 70.7\% resolution
with eleven test-time attempts.
Recovery with deployment-time verifiers remains future work.

%% file: appendix.tex
\section{Verifier Choice and Evaluation Protocol}
\label{app:verifier}

\paragraph{A common verification procedure.}
Our study concerns learning to recover from failed attempts, rather
than designing a new verifier. We use the official SWE-bench
verification procedure as the common execution-based verifier during
training and evaluation~\citep{jimenez2023swebench}. It is applied to
the task-specific tests of the respective datasets: R2E-Gym for
training and SWE-bench Verified for evaluation. Each submitted patch
receives a binary outcome and diagnostic output that can condition a
subsequent attempt. Sharing this verification procedure does not mean
sharing training and evaluation tasks.

\paragraph{Why we do not introduce a separate verifier.}
Verification and recovery-policy learning are related but distinct
problems. R2E-Gym combines execution-based and execution-free
verification~\citep{jain2025r2egym}. DeepSWE uses learned trajectory
scoring and generated tests for candidate selection, and evaluates
selected patches with the official SWE-bench harness~\citep{deepswe2025}.
Our question is instead how a policy learns to produce a better
attempt after observing evidence of failure. Introducing a new
verifier would vary both the policy-training method and its feedback
source. We therefore keep the verification procedure, feedback
construction, and evaluation attempt budgets the same across Base,
GRPO, and \method{} in the controlled Qwen3.5-4B comparison, except
where feedback content is explicitly ablated. This isolates the
training comparison from changes to the verifier; it does not imply
that the official verifier is optimal for deployment.

\paragraph{What the evaluation measures.}
The controller observes benchmark-test outcomes and diagnostic
feedback between attempts and stops after success. Multi-attempt
results therefore measure \emph{verifier-assisted recovery}, not
independent-sampling Pass@$K$ or candidate selection without access to
benchmark-test feedback. Using the official success criterion does
not make this feedback access identical to that of other published
systems. All controlled comparisons in this paper use the same
access conditions.

\section{Detailed Experimental Protocol}
\label{app:details}

\subsection{Data, Initialization, and Checkpoint Selection}

Training uses 4{,}518 tasks from ten R2E-Gym
repositories~\citep{jain2025r2egym}, with no repository overlap with
SWE-bench Verified. GRPO and \method{} are separately initialized from
the same Qwen3.5-4B checkpoint. OpenHands CodeActAgent generates
training trajectories, whereas evaluation uses
SWE-agent~\citep{wang2024openhands,yang2024sweagent}.
The reported FC-SWE checkpoint (step 67) was selected based on
performance across multiple validation sets, including a fixed
100-task subset of SWE-bench Verified.
RL parameter updates used the R2E-Gym training data.

Training was resumed across scheduler allocations. Curves are
stitched by global optimization step, not by each allocation's local
index. The canonical segments cover steps 1--36, 37--50, 51--67, and
68--84. The reward-sharing and penalty-removal ablations continue
from the same step-26 checkpoint; their comparison concerns the
subsequent training segment (Section~\ref{sec:dynamics}).

\subsection{Hyperparameters}

Table~\ref{tab:hyperparameters} summarizes the main configuration.
Sixteen tasks with eight initial chains each yield 128 initial
attempts per update. Up to 128 additional attempts are activated by
initial failures; unused positions in the 256-row transport tensor
are padding, not sampled trajectories.

\begin{table}[t]
\caption{Main training and evaluation configuration.}
\label{tab:hyperparameters}
\centering\small
\setlength{\tabcolsep}{4pt}
\begin{tabular}{@{}p{0.39\linewidth}p{0.57\linewidth}@{}}
\toprule
Item & Value \\
\midrule
Policy initialization & Qwen3.5-4B \\
Training / evaluation harness & OpenHands CodeActAgent / SWE-agent \\
Context budget & 65{,}536 tokens \\
Maximum model--tool interactions & 100 per attempt \\
Training attempt budget & $R=2$ \\
Tasks / initial chains per task & 16 / 8 \\
Physical rollout rows & 256, including inactive padding \\
Learning rate / schedule & $10^{-6}$, constant, no warmup \\
Optimizer & Adam; $\beta_1=0.9$, $\beta_2=0.999$, $\epsilon=10^{-8}$; weight decay 0 \\
Policy clipping & $\epsilon_l=0.2$, $\epsilon_h=0.28$ \\
Gradient-norm clipping & 1.0 \\
Reference-model KL & None \\
Loss normalization & Total valid generated tokens across the distributed batch \\
Runtime-local penalties & Token advantages overwritten with $-5$ in assistant messages flagged for invalid tool calls or malformed thinking \\
Agent / command timeout & 1{,}800 s / 120 s \\
Training / rollout parallelism & TP4, PP2, CP2 / TP1 \\
Evaluation decoding & Temperature 0.6; top-$p$ 0.95; top-$k$ 20 \\
Decoding seeds & 2, 3, 4 unless otherwise specified \\
Failure-evidence limits & 30{,}000-token carry-over; 6{,}000-character verifier excerpt; 3{,}000-character failure snippet \\
Training evidence construction & Same as the evaluation controller \\
\bottomrule
\end{tabular}
\end{table}

\subsection{Recovery Context and Trajectory Bookkeeping}

The controller transformation $\mathcal{C}$ in
Section~\ref{sec:chains} constructs recovery context in a fixed order.
It states that the previous attempt failed and that the new attempt
starts from a clean repository. It then prioritizes diagnostic
evidence---tracebacks, assertion failures, syntax or import errors,
explicit test failures, and timeouts---and appends deduplicated
verifier context within the excerpt budget. The previous candidate
patch is included subject to middle truncation and the carry-over
budget. The prompt explicitly permits keeping, revising, or
discarding the previous solution. The budgets in
Table~\ref{tab:hyperparameters} apply during both training and evaluation.

Each rollout row retains a task identifier derived from its dataset
and instance ID. This identifier determines the active-set comparison
group in Section~\ref{sec:advantages}; a separate chain identifier
tracks attempts and per-chain metrics. Inactive transport rows
retain bookkeeping identifiers but enter neither reward statistics
nor the policy loss. Copied failure evidence is input context and
does not receive policy-loss weight.

\subsection{Serialization and Runtime Consistency}

Long tool-use trajectories require consistent message boundaries
across tokenization, serving, the agent harness, and stored rollouts.
The implementation uses a fixed inline serving template and the
model family's reasoning and tool-call parsers. It preserves
assistant reasoning content, renders assistant text and structured
tool calls deterministically, groups tool observations under the
expected message boundaries, and aligns the generation prefix and
thinking flag between tokenizer and server. Invalid tool calls are
identified from runtime execution results rather than an additional
text-schema check. Prefix caching is disabled in the formal runs as
an implementation setting, not as a requirement of \method{}.

\subsection{Metrics and Seed Aggregation}
\label{app:metrics}

Let $z_{s,n,k}$ indicate that task $n$ is first resolved on attempt
$k$ under decoding seed $s$, with $N=500$ tasks and $S$ seeds. We report
percentages:
\begin{align}
\mathrm{Resolved@}K
&=\frac{100}{SN}\sum_{s=1}^{S}\sum_{n=1}^{N}
  \mathbb{1}\!\left[\sum_{k=1}^{K}z_{s,n,k}>0\right],\\
\mathrm{ConditionalRecovery@}K
&=100\,
\frac{\sum_{s,n}\mathbb{1}[z_{s,n,1}=0]
       \mathbb{1}[\sum_{k=2}^{K}z_{s,n,k}>0]}
     {\sum_{s,n}\mathbb{1}[z_{s,n,1}=0]}.
\end{align}
Resolved@1 is first-attempt success. Because every seed evaluates
the same number of tasks, the first expression also equals the mean
of seed-level resolution rates. Conditional recovery instead pools
recovered tasks and initial failures across seeds; it is not the
unweighted mean of seed-level recovery percentages.

Conditional recovery is an incorrect-to-correct measure conditioned
on initial failure~\citep{kumar2024score}. Success terminates the
chain, so the protocol does not generate subsequent attempts from
already-correct patches. The three seeds are \emph{decoding seeds},
not three independent RL training runs. Table~\ref{tab:ablations},
block B, and Table~\ref{tab:share-ablation} use one decoding seed.
Full per-seed scaling counts appear in Table~\ref{tab:scaling-full}.

\section{Case Studies: Revising a Failed Patch and Preserving Existing Behavior}
\label{app:case-study}

\begin{figure}[!htb]
\centering
\includegraphics[width=\linewidth]{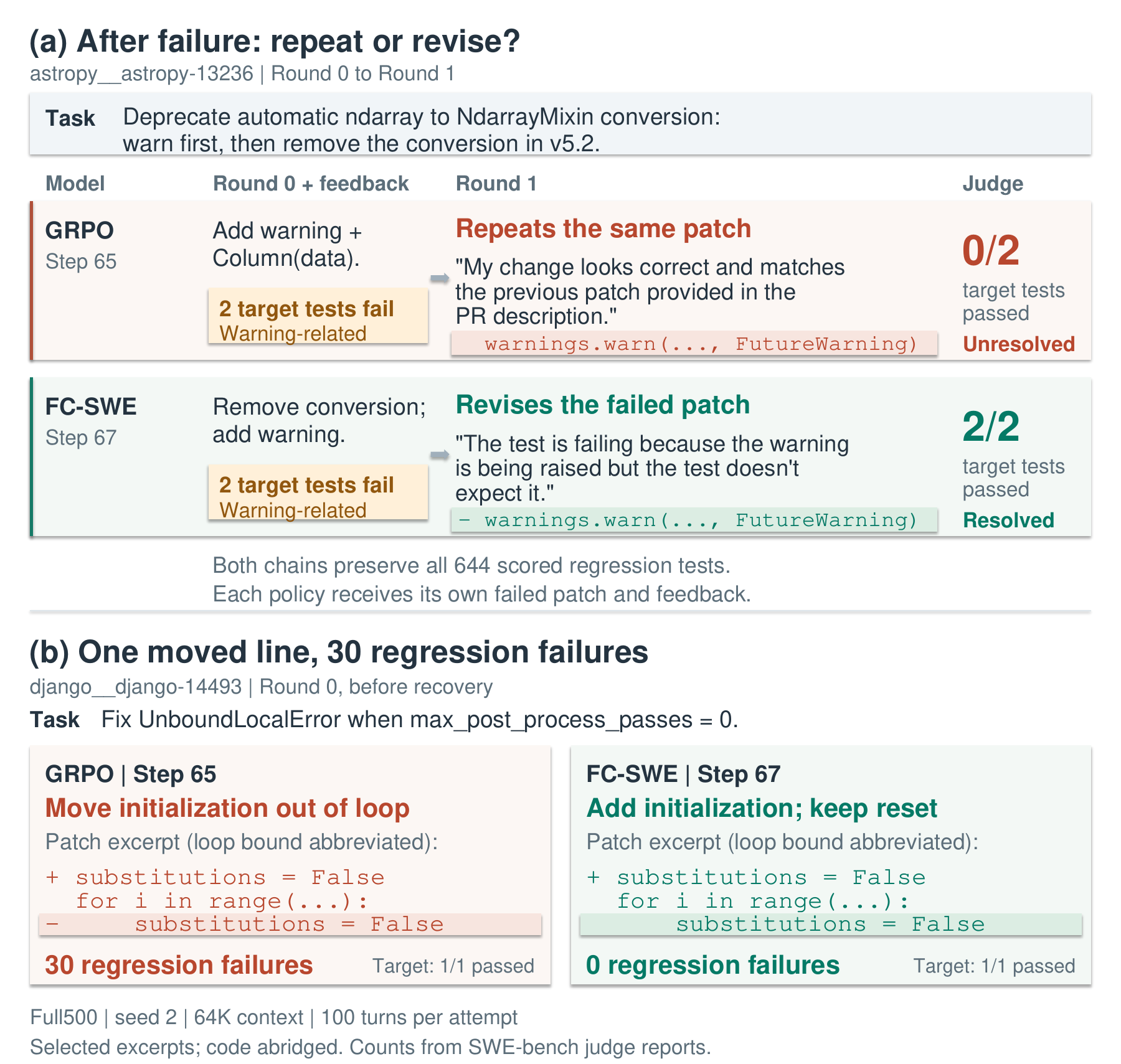}
\caption{Selected SWE-bench Verified cases comparing GRPO and
\method{}. (a) Second-attempt responses to each policy's own failed
patch and verifier feedback. (b) First-attempt patches that both
pass the target test but differ in regression-test failures.}
\label{fig:case-studies}
\end{figure}

\paragraph{How to read the examples.}
A \emph{patch} is the set of code edits submitted by an agent.
\emph{Target tests} check the reported problem, whereas
\emph{regression tests} check that previously working behavior still
works. Passing a target test is therefore insufficient if the patch
breaks existing functionality. Figure~\ref{fig:case-studies} compares
GRPO at step 65 with the final \method{} policy at step 67, using
decoding seed 2, a 64K context, and at most 100 interactions per
attempt. The figure's Round 0 and Round 1 denote the first and second
attempts, respectively. These are selected examples rather than an
estimate of how frequently either behavior occurs; code fragments
are abridged and test counts refer to the scored tests.

\subsection{Revising a Failed Attempt After Verifier Feedback}
\label{app:case-recovery}

\paragraph{The task in plain language.}
In \texttt{astropy\_\_astropy-13236}, adding a structured numerical
array to an Astropy table automatically converts it into a special
wrapper, \texttt{NdarrayMixin}. The issue discusses moving away from
this conversion, initially warning users and later changing the
default behavior. The challenge is to change how the table handles
the array without introducing incompatible behavior elsewhere.

\paragraph{Initial failure and subsequent revision.}
In Figure~\ref{fig:case-studies}(a), both initial patches fail the
two scored target tests, with warning-related diagnostic feedback.
GRPO adds a warning and a \texttt{Column(data)} conversion, then
repeats its patch on the second attempt; both target tests still
fail. \method{} initially removes the automatic conversion and adds
a warning. On its second attempt it removes the warning, after
which both target tests pass. Both policies preserve all 644 scored
regression tests in this example.

\paragraph{What this example shows.}
This example contrasts repeating a rejected patch with revising it
following feedback. Each policy receives its \emph{own} failed patch
and feedback after a clean reset, so the starting patches differ.
Passing the benchmark tests does not establish full compliance with
the proposed deprecation behavior; the example illustrates
feedback-conditioned revision, not a general benefit from removing
warnings.

\subsection{Fixing the Reported Error Without Introducing Regressions}
\label{app:case-regression}

\paragraph{The task in plain language.}
In \texttt{django\_\_django-14493}, a file-processing routine crashes
when its maximum number of processing passes is zero. A flag named
\texttt{substitutions} is initialized only inside the loop, so it
has no value if the loop never runs. The flag must be initialized
before the loop while retaining the reset needed during each
normal processing pass.

\paragraph{Two superficially similar edits.}
Figure~\ref{fig:case-studies}(b) shows that GRPO moves the
initialization outside the loop and removes its per-pass reset.
\method{} instead adds an initialization before the loop while
retaining the reset inside it. Both initial patches pass the one
target test, but GRPO's patch causes 30 regression-test failures,
whereas the \method{} patch causes none. Thus, the same apparent
fix to the reported error can have different consequences for
existing behavior.

\paragraph{What this example shows.}
This contrast occurs on the \emph{first attempt}, before either
policy receives post-verification recovery feedback. It illustrates
patch quality and the role of regression tests, not a successful
second-attempt recovery. Together, the examples distinguish revising
a failed solution from preserving correctness outside the reported
failure. Aggregate results, rather than these selected examples,
support the comparisons in Section~\ref{sec:experiments}.
\section{Additional Ablation Results}
\label{app:ablations}

\subsection{What the Ablations Isolate}

The training-time failure-context ablation retains failure-triggered
retries, clean resets, trajectory-local rewards, and active-set
advantage estimation, but removes the failed patch and verifier
feedback from recovery prompts during training. Both trained
policies are evaluated with failure context. This separates the
effect of conditioning training on failure evidence from simply
sampling additional attempts (Table~\ref{tab:ablations}, block A).

The reward-sharing ablation broadcasts the final chain outcome to
all executed trajectories, turning labels $[0,1]$ into $[1,1]$.
The additional penalty-removal ablation changes runtime-local
penalties \emph{within this shared-reward configuration}. It does not
isolate the penalty effect under trajectory-local rewards.
Alternative advantage-group constructions and loss normalizations
are motivated in Section~\ref{sec:method}, but are not separately
ablated. The experiments therefore do not quantify the independent
contribution of every implementation choice.

\subsection{Shared-Reward Checkpoint Analysis}

Table~\ref{tab:share-ablation} reports single-seed evaluations on all
500 tasks. The shared-reward run improves across its evaluated
checkpoints, so reward sharing does not prevent learning. At the
checkpoints used in Table~\ref{tab:ablations}, trajectory-local
rewards give 43.2\% Resolved@1 and 54.8\% Resolved@2, compared with
41.4\% and 51.8\% for shared rewards. The 3.0-percentage-point
Resolved@2 gap consists of 1.8 points in first-attempt resolution
and 1.2 points in additional second-attempt resolutions; it is not
entirely attributable to recovery. Conditional recovery is 20.4\%
versus 17.7\%. These are comparisons of selected checkpoints at
different steps, not a step-matched learning curve.

\begin{table}[htbp]
\caption{Checkpoint evaluations on all 500 tasks using one decoding
seed. Recovered counts tasks first resolved on attempt 2.}
\label{tab:share-ablation}
\centering\small
\setlength{\tabcolsep}{3pt}
\begin{tabular*}{\linewidth}{@{\extracolsep{\fill}}lrrrr@{}}
\toprule
Checkpoint & Resolved@1 & Recovered & Recovery & Resolved@2 \\
\midrule
Shared, step 40 & 37.6\% & 41 & 13.1\% & 45.8\% \\
Shared, step 45 & 39.4\% & 41 & 13.5\% & 47.6\% \\
Shared, step 55 & 42.2\% & 47 & 16.3\% & 51.6\% \\
Shared, step 70 & 41.4\% & 52 & 17.7\% & 51.8\% \\
\midrule
Local, step 50 & 42.4\% & 53 & 18.4\% & 53.0\% \\
Local, step 67 & 43.2\% & 58 & 20.4\% & 54.8\% \\
Local, step 84 & 43.8\% & 44 & 15.7\% & 52.6\% \\
\bottomrule
\end{tabular*}
\par\smallskip
\begin{minipage}{\linewidth}\footnotesize
``Recovery'' denotes conditional recovery among first-attempt failures.
``Local'' denotes trajectory-local rewards in \method{}.
\end{minipage}
\end{table}

The local-reward run's first-attempt resolution increases between
steps 67 and 84, but its recovery rate and Resolved@2 decrease.
Later training therefore does not uniformly improve both
capabilities. With runtime-local penalties also removed from the
shared-reward configuration, online success falls from 76.6\% at
step 28 to 2.3\% at step 31 in the observed post-checkpoint segment.
This run has no downstream evaluation; a missing table entry does
not mean zero benchmark resolution.

\subsection{Additional Feedback-Content Controls}

For the final \method{} policy, verifier-only context reaches 20.4\%
conditional recovery, compared with 19.0\% for verifier evidence
plus the failed patch, 15.7\% for patch-only context, and 17.0\%
without failure context. The point estimates support the usefulness
of the verifier channel, but do not show that including the patch
is necessary or that the default combined context is optimal.
We do not infer statistical superiority between verifier-only and
combined context from these percentages alone.

A same-task shuffled-evidence control is also inconclusive about
the importance of pairing feedback with a particular failed patch.
Different failed attempts on the same task can expose the same
failing tests, so this shuffle can retain useful task-level
information. It is not equivalent to removing verifier evidence.
The main claims rely on the directly reported context contrasts in
Figure~\ref{fig:feedback-mechanism}, together with the training-time
failure-context ablation.

\section{Additional Test-Time Scaling and Cost Accounting}
\label{app:scaling-details}

\subsection{Per-Seed Resolution Counts}

\begin{figure}[t]
\centering
\includegraphics[width=\linewidth]{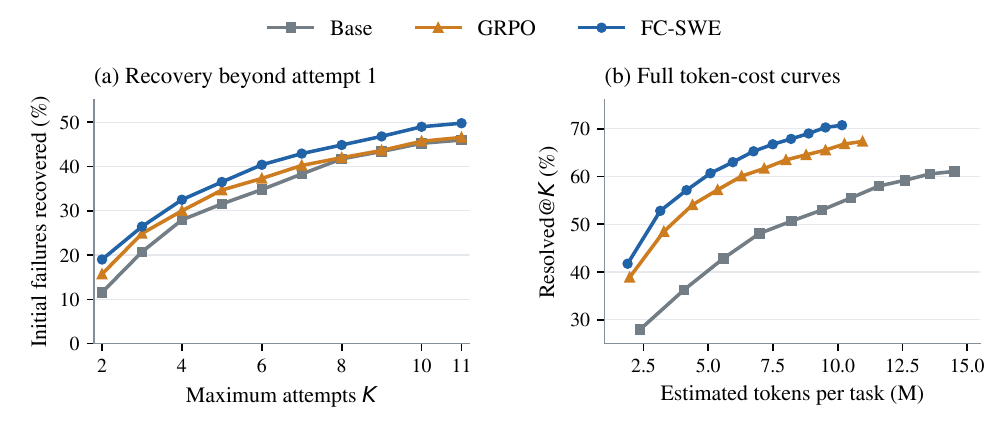}
\caption{Additional test-time scaling views.
(a) Initial failures recovered within $K$ attempts, pooled across
seeds. (b) Cumulative resolution versus estimated token cost,
reproduced for reference; markers denote evaluated attempt budgets.}
\label{fig:scaling-cost-details}
\label{fig:scaling}
\label{fig:cost-frontier}
\end{figure}

Table~\ref{tab:scaling-full} gives the cumulative solved counts behind
Figure~\ref{fig:scaling-efficiency}(a). The same policy is used at all
test-time attempt budgets; training uses at most two attempts.
Figure~\ref{fig:scaling-cost-details}(a) expresses these results as
the fraction of first-attempt failures recovered by budget $K$.

\begin{table}[htbp]
\caption{Cumulative solved tasks out of 500. Each cell gives counts
for decoding seeds 2/3/4, followed by the mean resolution rate.}
\label{tab:scaling-full}
\centering\small
\setlength{\tabcolsep}{4pt}
\begin{tabular*}{\linewidth}{@{\extracolsep{\fill}}rccc@{}}
\toprule
$K$ & Base & GRPO & \method{} \\
\midrule
1 & 150/124/145 (27.9\%) & 206/179/198 (38.9\%) & 216/207/203 (41.7\%) \\
2 & 183/168/193 (36.3\%) & 257/228/242 (48.5\%) & 274/261/257 (52.8\%) \\
3 & 221/193/229 (42.9\%) & 278/255/278 (54.1\%) & 289/280/288 (57.1\%) \\
4 & 241/227/253 (48.1\%) & 292/269/297 (57.2\%) & 305/296/309 (60.7\%) \\
5 & 251/243/266 (50.7\%) & 304/284/313 (60.1\%) & 318/306/321 (63.0\%) \\
6 & 260/258/277 (53.0\%) & 310/298/317 (61.7\%) & 329/320/330 (65.3\%) \\
7 & 268/270/295 (55.5\%) & 317/305/330 (63.5\%) & 337/327/337 (66.7\%) \\
8 & 280/283/307 (58.0\%) & 323/311/334 (64.5\%) & 344/331/343 (67.9\%) \\
9 & 286/289/313 (59.2\%) & 328/314/341 (65.5\%) & 352/335/348 (69.0\%) \\
10 & 297/294/317 (60.5\%) & 336/322/344 (66.8\%) & 359/338/357 (70.3\%) \\
11 & 299/297/320 (61.1\%) & 340/325/345 (67.3\%) & 362/341/358 (70.7\%) \\
\bottomrule
\end{tabular*}
\end{table}

Taking the union of solved tasks across the three seeded chains
gives 78.8\% for \method{}, 75.8\% for GRPO, and 73.2\% for Base at
$K=11$. This union permits up to three times as many attempts per
task and is not used for the matched-budget comparisons.

\subsection{Token Accounting and Coverage}

For each decoding seed, we count all executed attempts through
budget $K$, excluding attempts after the task's first success. We
sum their costs over all 500 tasks, including unresolved tasks,
divide by 500, and average the three seed-level means.
Estimated token cost combines recorded input-token counts with
generated tokens estimated from output characters using a
calibrated ratio of 4.0964 characters per token. Repeated processing
of input context at successive model calls is included. This differs
from the output-only generation length in
Appendix~\ref{app:execution-diagnostics}.

Three Base second-attempt trajectory logs are missing for seed 2.
Their token costs are omitted, undercounting Base's cost; the
corresponding resolution outcomes remain available. The token-cost
comparison is therefore approximate. It is neither a direct FLOP
measurement nor measured end-to-end latency, and a matched attempt
budget does not imply identical token usage.

\subsection{Observed Resolution--Cost Operating Points}

Figure~\ref{fig:scaling-efficiency}(b) plots the full observed
resolution--token-cost curves, with one point per attempt budget
$K=1,\ldots,11$. The target-resolution comparisons in the text select
the lowest-cost evaluated budget reaching \emph{at least} the
target. For 50\%, the selected budgets are $K=5,3,2$ for Base, GRPO,
and \method{}, at 8.22M, 4.39M, and 3.15M tokens per task. For 60\%,
they are $K=10,5,4$, at 13.57M, 6.29M, and 5.08M tokens.
These are observed discrete points, not interpolated equal-score
operating points. Their attained resolution can differ, so these
comparisons do not establish an exact equal-accuracy speedup.

\subsection{Full Per-Budget Accounting}

Table~\ref{tab:per-k-cost-full} reports token costs, executed attempts,
and cumulative resolution. Each executed attempt produces one
benchmark verification, so verifier-call counts equal attempt
counts. The auxiliary time estimates combine measured tool-execution
time with serving-throughput rates of 15{,}500 prefill and 145 decode
tokens per second. They are synthesized processing-time estimates
for 500 tasks, not observed end-to-end wall-clock times; parallel
scheduling, infrastructure overhead, and verifier execution are not
fully characterized by this estimate. We do not use this column to
claim an end-to-end speedup.

\begin{table}[!htbp]
\caption{Per-budget inference accounting, averaged over three
decoding seeds. Token costs are estimated; the hours column is a
synthesized estimate for all 500 tasks, not measured latency.}
\label{tab:per-k-cost-full}
\centering\small
\setlength{\tabcolsep}{4pt}
\begin{tabular*}{\linewidth}{@{\extracolsep{\fill}}rrrrr@{}}
\toprule
$K$ & Tokens/task (M) & Attempts/task & Synth. hours & Resolved@$K$ \\
\midrule
\multicolumn{5}{@{}l}{\textit{Base}} \\
1 & 2.36 & 1.00 & 33 & 27.9\% \\
2 & 4.06 & 1.72 & 57 & 36.3\% \\
3 & 5.59 & 2.36 & 78 & 42.9\% \\
4 & 6.98 & 2.93 & 96 & 48.1\% \\
5 & 8.22 & 3.45 & 113 & 50.7\% \\
6 & 9.40 & 3.94 & 129 & 53.0\% \\
7 & 10.52 & 4.41 & 145 & 55.5\% \\
8 & 11.60 & 4.85 & 159 & 58.0\% \\
9 & 12.60 & 5.27 & 173 & 59.2\% \\
10 & 13.57 & 5.68 & 186 & 60.5\% \\
11 & 14.51 & 6.08 & 199 & 61.1\% \\
\midrule
\multicolumn{5}{@{}l}{\textit{GRPO}} \\
1 & 1.96 & 1.00 & 28 & 38.9\% \\
2 & 3.29 & 1.61 & 47 & 48.5\% \\
3 & 4.39 & 2.13 & 63 & 54.1\% \\
4 & 5.36 & 2.59 & 77 & 57.2\% \\
5 & 6.29 & 3.01 & 90 & 60.1\% \\
6 & 7.16 & 3.41 & 102 & 61.7\% \\
7 & 8.00 & 3.80 & 114 & 63.5\% \\
8 & 8.78 & 4.16 & 125 & 64.5\% \\
9 & 9.53 & 4.52 & 136 & 65.5\% \\
10 & 10.26 & 4.86 & 146 & 66.8\% \\
11 & 10.96 & 5.19 & 156 & 67.3\% \\
\midrule
\multicolumn{5}{@{}l}{\textit{FC-SWE}} \\
1 & 1.89 & 1.00 & 28 & 41.7\% \\
2 & 3.15 & 1.58 & 46 & 52.8\% \\
3 & 4.16 & 2.05 & 60 & 57.1\% \\
4 & 5.08 & 2.48 & 73 & 60.7\% \\
5 & 5.95 & 2.88 & 85 & 63.0\% \\
6 & 6.75 & 3.25 & 97 & 65.3\% \\
7 & 7.49 & 3.59 & 107 & 66.7\% \\
8 & 8.20 & 3.93 & 117 & 67.9\% \\
9 & 8.88 & 4.25 & 126 & 69.0\% \\
10 & 9.53 & 4.56 & 136 & 70.3\% \\
11 & 10.17 & 4.86 & 144 & 70.7\% \\
\bottomrule
\end{tabular*}
\end{table}

\clearpage
\section{Additional Training Diagnostics}
\label{app:execution-diagnostics}

\subsection{Definitions and Averaging Population}

\begin{figure}[!htbp]
\centering
\includegraphics[width=\linewidth]{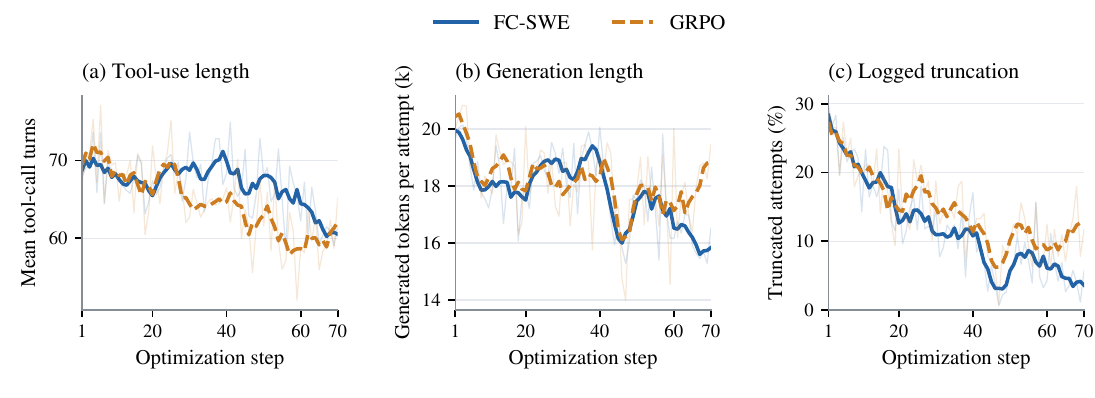}
\caption{Training execution statistics per active attempt:
(a) tool-call turns, (b) generated tokens, and (c) logged truncation.
Light lines show individual steps; dark lines show centered
five-step averages with shortened endpoint windows.}
\label{fig:execution-diagnostics}
\end{figure}

Figure~\ref{fig:execution-diagnostics} reports tool-use length,
generation length, and logged truncation during training. For
\method{}, each per-step statistic includes all active initial and
recovery attempts, weighted by their actual counts; unexecuted
recovery slots are excluded. GRPO contributes initial attempts only.
These are individual training histories, not means over the three
decoding seeds used for benchmark evaluation. Their attempt
populations differ and change as the policies improve.

Tool-call turns count assistant messages flagged as tool calls,
including invalid calls. Generation length sums assistant-generated
tokens within an attempt, excluding input and tool-observation
tokens. Logged truncation indicates that recorded total trajectory
tokens reach the configured token budget. It is not the fraction
reaching the 100-interaction limit and does not distinguish all
termination reasons. Mean generated length and truncation measure
different quantities: the former is an output-token average,
whereas the latter is a budget-threshold event for total trajectory
tokens. Similar generation-length means therefore need not imply
similar truncation rates.

\subsection{Observed Trends and Interpretation}

Over steps 61--70, \method{} averages 62.0 tool-call turns and 16.2k
generated tokens per active attempt, compared with 60.4 turns and
17.8k tokens for GRPO. Logged truncation rates are 5.2\% and 10.5\%,
respectively. These describe execution behavior, not total training
cost: \method{} also generates recovery attempts. Lower truncation
does not by itself imply successful termination or establish the
cause of improved test-time efficiency. Test-time token efficiency
is assessed separately in Section~\ref{sec:beyond}.

\paragraph{Runtime settings for cross-model reference results.}
The controlled Qwen3.5-4B policies use a 30-minute agent timeout.
The Qwen3.5-9B and Nemotron-3-Nano-4B reference results use a
60-minute timeout. Those rows are therefore not timeout-matched to
the trained Qwen3.5-4B policies and are not used to isolate the
effect of recovery training.

\clearpage
\section{Limitations and Future Work}
\label{app:future-work}

\paragraph{Recovery with practical verifiers.}
The evaluation assumes access to benchmark-test feedback after
each attempt. A next step is to use verifiers available during
deployment, such as learned critics, agent-generated tests, or
hybrid verification~\citep{pan2024swegym,jain2025r2egym,deepswe2025}.
This requires separating the verifier that guides recovery from
held-out tests used for final scoring. Evaluation should measure
verifier cost, false acceptance, false rejection, and diagnostic
usefulness: detecting an incorrect patch and explaining how to
repair it are distinct requirements.

\paragraph{Imperfect feedback and incomplete specifications.}
Passing a finite test suite does not guarantee semantic correctness
or complete satisfaction of an issue description. Recovery training
should be evaluated when feedback is incomplete, noisy, or
misleading, including whether the policy checks evidence rather than
following it uncritically. Cross-verifier evaluation and independent
patch review would help distinguish robust repair from adaptation
to a particular feedback source.

\paragraph{Experimental coverage.}
The controlled training study uses Qwen3.5-4B and one training
configuration. Multiple decoding seeds characterize inference
variation, not variation across independently trained policies.
Some ablations have only one decoding seed, and the penalty-removal
comparison is confined to shared-reward training. Extending the
study to multiple training seeds, model sizes, task distributions,
and separately controlled optimization components would test the
generality of the findings. The selected cases illustrate behaviors
but do not establish their prevalence.

\paragraph{Compute-matched training and deployment cost.}
\method{} generates additional trajectories during training, so the
matched-task, matched-initial-group comparison is not matched in
total training compute. Future studies should examine quality under
equal training-token or wall-clock budgets. At inference, exact
tokenization, complete trajectory logs, verifier execution cost,
and measured latency would complement the present approximate
token-accounting analysis. These extensions are needed before
claiming end-to-end deployment savings.